\pdfoutput=1
\documentclass[11pt]{article}

\usepackage{emnlp2021}

\usepackage{times}
\usepackage{latexsym}

\usepackage[T1]{fontenc}
\usepackage[utf8]{inputenc}

\usepackage{microtype}

\usepackage{times}  % DO NOT CHANGE THIS
\usepackage{helvet} % DO NOT CHANGE THIS
\usepackage{courier}  % DO NOT CHANGE THIS
\usepackage{graphicx} % DO NOT CHANGE THIS
\usepackage{subcaption}
\usepackage{natbib}  % DO NOT CHANGE THIS AND DO NOT ADD ANY OPTIONS TO IT
\usepackage{caption} % DO NOT CHANGE THIS AND DO NOT ADD ANY OPTIONS TO IT
\usepackage{amsmath}
\usepackage{booktabs}
\usepackage{xcolor}
\usepackage{xspace}
\usepackage{enumitem}
\usepackage{multirow, multicol}
\usepackage{flushend}

\usepackage{amsfonts}
\newcommand{\abr}[1]{\textsc{#1}}
\newcommand{\dn}{\abr{Comet}\xspace} % dataset name - TBD

\newcommand{\reffig}[1]{Fig.~\ref{#1}}
\newcommand{\refsec}[1]{Sec.~\ref{#1}}
\newcommand{\reftab}[1]{Tab.~\ref{#1}}

\newcommand{\reportval}[2]{$#1$\footnotesize $\pm #2$}

\usepackage{booktabs,arydshln}
\makeatletter
\def\adl@drawiv#1#2#3{%
        \hskip.5\tabcolsep
        \xleaders#3{#2.5\@tempdimb #1{1}#2.5\@tempdimb}%
                #2\z@ plus1fil minus1fil\relax
        \hskip.5\tabcolsep}
\newcommand{\cdashlinelr}[1]{%
  \noalign{\vskip\aboverulesep
           \global\let\@dashdrawstore\adl@draw
           \global\let\adl@draw\adl@drawiv}
  \cdashline{#1}
  \noalign{\global\let\adl@draw\@dashdrawstore
           \vskip\belowrulesep}}
\makeatother

\belowdisplayskip \abovedisplayskip

\newlength{\sectionReduceTop}
\newlength{\sectionReduceBot}
\newlength{\subsectionReduceTop}
\newlength{\subsectionReduceBot}
\newlength{\abstractReduceTop}
\newlength{\abstractReduceBot}
\newlength{\captionReduceTop}
\newlength{\captionReduceBot}
\newlength{\subsubsectionReduceTop}
\newlength{\subsubsectionReduceBot}

\newlength{\eqnReduceTop}
\newlength{\eqnReduceBot}

\newlength{\horSkip}
\newlength{\verSkip}

\newlength{\figureHeight}
\usepackage{graphicx} 
\usepackage{multicol}
\usepackage{booktabs} 
\usepackage{algorithm} 
\usepackage{algorithmic} 
\usepackage{color}
\usepackage{amsmath}
\usepackage{amssymb}
\usepackage{amsfonts}
\usepackage{multirow, varwidth}
\usepackage{stfloats} 
\usepackage{verbatim}
\makeatletter
\newif\if@restonecol
\makeatother
\usepackage{xr}
\usepackage[amsmath,thmmarks]{ntheorem}

\usepackage{xspace}
\makeatletter
\DeclareRobustCommand\onedot{\futurelet\@let@token\@onedot}
\def\@onedot{\ifx\@let@token.\else.\null\fi\xspace}

\def\eg{\emph{e.g}\onedot} 
\def\ie{\emph{i.e}\onedot}

\def\etc{\emph{etc}\onedot}

\makeatother

\newcommand{\sm}[1]{{\color{black}{#1}}}
\newcommand{\emptydraft}[1]{{\color{black}{}}}

\newcommand{\sk}[1]{{\color{black}{#1}}}
\usepackage[misc]{ifsym} % Letter
\newcommand*\samethanks[1][\value{footnote}]{\footnotemark[#1]}

\title{Navigating Connected Memories with a Task-oriented Dialog System}

\author{ \\
  \textbf{Seungwhan Moon}\thanks{\hspace{5pt}Joint first authors},
  \textbf{Satwik Kottur}\samethanks,
  \textbf{Alborz Geramifard},
  \textbf{Babak Damavandi} \\
  Meta Reality Labs \& Meta AI \\
  {\small \Letter}\hspace{3pt} \texttt{\{shanemoon,skottur,alborzg,babakd\}@fb.com} \\
}
\date{}

\begin{document}
\maketitle
\begin{abstract}
     Recent years have seen an increasing trend in the volume of personal media captured by users, thanks 
    to the advent of smartphones and smart glasses, resulting in large media collections.
    Despite conversation being an intuitive human-computer interface, current efforts focus mostly
    on single-shot natural language based media retrieval to aid users query their media and 
    re-live their memories. This severely limits the search functionality as users can neither ask 
    follow-up queries nor obtain information without first formulating a single-turn query.

    In this work, we propose \textit{dialogs for connected memories} as a powerful tool to empower
    users to search their media collection through a multi-turn, interactive conversation.
    Towards this, we collect a new task-oriented dialog dataset \dn, which contains $11.5k$ 
    user$\leftrightarrow$assistant dialogs (totalling $103k$ utterances), grounded in 
    simulated personal memory graphs.
    We employ a resource-efficient, two-phase data collection pipeline that uses:
    (1) a novel multimodal dialog simulator that generates synthetic dialog flows grounded in 
    memory graphs, and,
    (2) manual paraphrasing to obtain natural language utterances.
    We analyze \dn, formulate four main tasks to benchmark meaningful progress, and adopt
    state-of-the-art language models as strong baselines, in order to highlight the 
    multimodal challenges captured by our dataset\footnote{Our code \& data is made available at \url{github.com/facebookresearch/comet_memory_dialog}}.
\end{abstract}

\section{Introduction}
\label{sec:introduction}

\begin{figure}[t!]
    \centering
    \includegraphics[width=0.92\columnwidth]{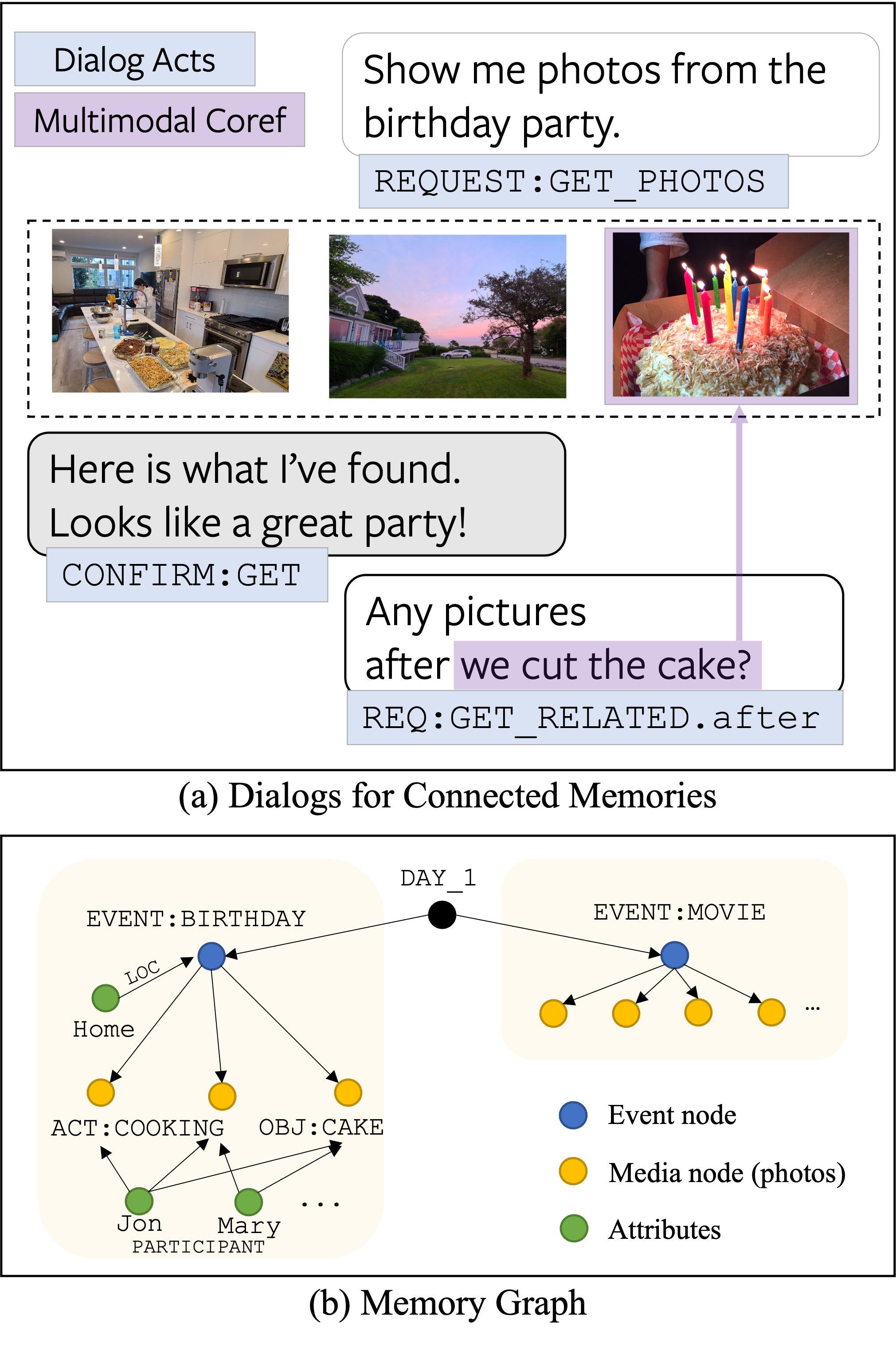}
    \vspace*{\captionReduceTop}
    \caption{Illustration of \dn: \textbf{CO}nnected \textbf{ME}mories with a \textbf{T}ask-oriented Dialog. (a) Each dialog turn is fully annotated with dialog acts and multimodal coreference labels, accompanied with photos associated with the request. (b) These media are from the underlying memory graph, a structured collection of personal media.}
    \vspace*{\captionReduceBot}
    \label{fig:teaser}
\end{figure}

\sk{
    The rise of smartphones and smart glasses has contributed to a surge in the amount of personal media 
    (photos, videos, montages, \etc) captured by users on a day-to-day basis in the past decade.
    For instance, it is estimated that about $1.5$ trillion photos would be clicked in the year $2022$ \cite{pantic_blogpost}.
    As a result, personal media collections typically grow at an alarming rate, making it 
    cumbersome for users to
    manually search, retrieve, and re-live their captured
    memories\footnote{Memories and media files are used interchangeably.}. % in this work
}

\sk{
    To alleviate this situation, solutions that perform natural language query-based media retrieval 
    \cite{drilldown,Vo_2019_CVPR,10.1145/1646396.1646442,DBLP:journals/corr/BarbuNS13,li2017person,dialog-image-retrieval,mmd} 
    have been proposed.
    However, such approaches exhibit two drawbacks.
    First, they are single-shot interactions without any context 
    carry-over, \eg, \textit{Show me some photos from the beach last week.}. 
    This limits the functionality and does not let users ask any follow-up queries like 
    \textit{`Display photos from the first time I was \underline{here}?'}, since understanding
    \underline{here} requires the query history.
    Second, users cannot seek information without actually formulating the query to retrieve the
    corresponding memory.
    For instance, there is no easy query to know the first time a user visited the beach in the memory
    they are reviewing.
}

\sk{
    In order to overcome these limitations, we propose \textit{dialogs for connected memories} 
    as a powerful interface where users can interactively query their memory collections.
    By design, a conversational agent can handle multi-turn interactions enabling several additional queries
    that require context carryover, \eg., \textit{`When was the first time I was at this beach?'}.
    Though prior efforts have explored the use of dialogs in media retrieval 
    \cite{fashion_iq,NEURIPS2018_a01a0380} in other domains
    (e.g., fashion), there is no existing work focusing on interactive search and query of
    personal media collections to the best of our knowledge.
}

\sk{
    More concretely, we propose \dn, a new multimodal task-oriented dialog dataset aimed at developing conversational
    assistants that can enable users to interactively search and query their collection of memories.
    Working with personal media collections presents two main obstacles:
    (a) There are no readily available public datasets that contain personal media along with 
    associated media attributes that we could leverage, and,
    (b) Personal memories constitute sensitive information, thus resulting in privacy and safety
    concerns.
    To circumvent these roadblocks, we devise a novel memory graph simulator that can leverage publicly
    available media datasets and help create several synthetic memory collections.
    We represent these collections as memory graphs to capture useful relationships between
    the constituent memories, \eg, memories taken at the same place.
    We then collect $11.5k$ user$\leftrightarrow$assistant task-oriented dialogs
    (totalling $103k$ utterances), grounded in $1.1k$ memory graphs.
    An example dialog is in \reffig{fig:teaser}.
}

\sk{
    Our dataset is challenging as it requires reasoning through both the dialog history
    and multimodal context (memory graphs) to resolve coreferences, track the dialog state, 
    predict the right API, and generate a meaningful natural language assistant response.
    As an example, consider the query \textit{`When was the first time I was \underline{here}?'}.
    First, the model needs to resolve \underline{here} using the dialog history and previously viewed
    memories.
    Next, it needs to understand that the query is seeking information about a \textit{connected memory},
    and predict the right API \texttt{get\_time(resolve(}\textit{here}\texttt{),} \textit{first time}\texttt{)}.
    Finally, it should produce a response like 
    \textit{`The first time you were here was on August 2, 2019 with Jean'}, potentially including
    some chit-chat.
}

\sk{
    To capture these challenges and benchmark progress towards assistants that can interactively 
    handle dialogs for connected memories, we formulate four main tasks: 
    Assistant API Call Prediction, Multimodal Coreference Resolution (MM-Coref), Multimodel
    Dialog State Tracking (MM-DST), and Response Generation.
    We train baseline models for these tasks, and discuss future research directions.
}

\section{Related Work}
\label{sec:related_work}

\noindent \textbf{Task-oriented Dialogs}
aim to understand user queries and accomplish a pre-defined set of tasks (\eg booking hotels), which is a popular setting in consumer-facing virtual assistants. 
Our work addresses similar challenges often found in other task-oriented dialogs, such as natural language understanding (NLU), dialog state tracking (DST) \cite{dstc2}, etc. 
Compared to the conventional task-oriented dialog datasets (\eg MultiWoZ \cite{multiwoz,multiwoz2.1,sgd-dst}), however, our work involves a unique multimodal setting where dialogs are grounded on a memory graph composed of several media files, introducing novel challenges such as Multimodal DST and Multimodal Coreference Resolution given personal photo collections.

The most notable modeling approaches for task-oriented dialog systems include casting the DST task as a joint causal language modeling problem \cite{simpletod,soloist,accentor}, by fine-tuning a large pre-trained transformers. %such as GPT-2 \cite{radford2019language}.
We follow this recent trend and provide baselines by extending it accommodate for the unique multimodal contexts that our dataset brings.

\noindent \textbf{Multimodal Dialogs}
have become increasingly more popular, where dialog models process both visual and text input to handle queries \cite{simmc2}.
Many existing literature \cite{avsd-dstc8,visdial,clevr-dialog,guesswhat,talk-the-walk,vision-dialog-navigation} study multimodal Q\&A dialogs grounded on a single image as grounding context, extending the conventional VQA \cite{vqa} tasks to multi-turn scenarios. 
\citet{xu-etal-2020-user} studies conversational recommendation system using personal memories.
We extend this line of work by studying the multimodal agent that operates on a collection of media (memory graph), thus requiring reasoning abilities over multiple grounding contexts.
Our focus on task-oriented dialogs extends the previous literature and datasets that primarily focus on retrieval tasks \cite{dialog-image-retrieval,mmd,firdaus2020multidm}, capturing structured user intents and fine-grained attributes for each multimodal query.

\noindent \textbf{Memory QA}: 
Our work is also similar to the Memory QA tasks \cite{memexqa, moon2019memory}, where the main task is to answer user QA queries upon a collection of images, extending the Visual QA task \cite{vqa} which operates on a single image.
However, the existing literature is limited to a simple single-turn QA interaction, and focuses on the identification of an evidential image to answer a question.
While our dataset does include QA queries, we extend the problem domain to the conversational settings which support complex scenarios (\eg searching for related memories), allowing for rich multimodal interactions.

%\noindent \textbf{Interactive Media Retrieval}
% already mentioned in intro, might skip here
%the conversational image retrieval tasks 

\section{The \dn Dataset}
\label{sec:dataset}
\begin{figure}[t]
    \centering
    \includegraphics[width=0.8\columnwidth]{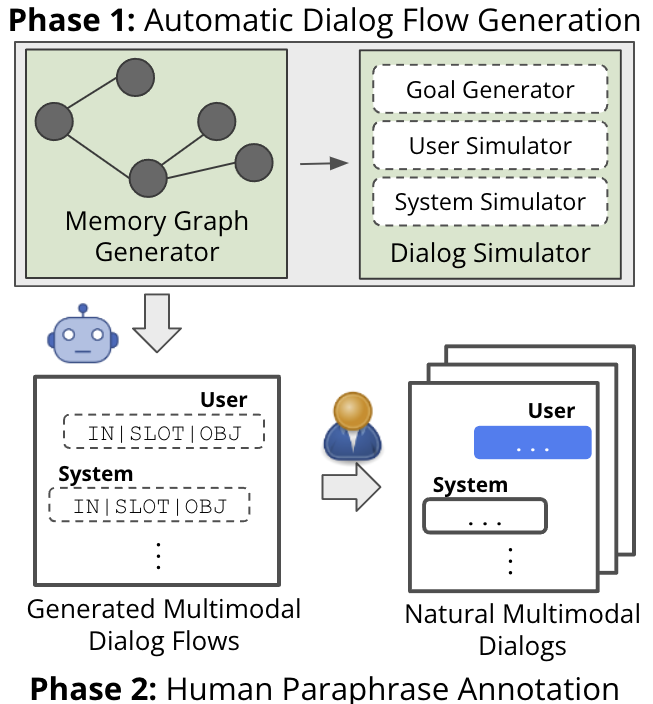}
    \caption{
        Two-stage pipeline to collect dialogs for \dn. 
        See \refsec{sec:dataset} for more details.
    }
    \vspace*{\captionReduceBot}
    \label{fig:data_collection_overview}
\end{figure}

\sk{
    \dn is aimed to enable assistant systems that can process interactive queries from users and
    help navigate their collection of memories through a natural language conversation.
    Towards this, we collect the \dn{} dataset using a two-phase approach (shown in \reffig{fig:data_collection_overview}):
    (a) Generating synthetic dialog flows between a user and an assistant that are conditioned on memory graphs, 
    using a novel multimodal dialog simulator (\refsec{sec:dataset_creation}), and,
    (b) Manually paraphrasing the above flows to obtain dialogs with natural language utterances 
    (\refsec{sec:human_paraphrase}), 
    thus moving closer to real-world application.
    %As argued in \cite{simmc2},
    This approach is resource-efficient as it reduces the annotation overheads when 
    compared to collecting human$\leftrightarrow$human dialogs, both in terms of cost and time.
    In what follows, we describe these two phases in detail and analyze our \dn dataset.
    See the supplementary (\reffig{fig:example}) for an example dialog.
}

\subsection{Multimodal Dialog Self-play}
\label{sec:dataset_creation}
\sk{
    %The first phase of our approach 
    We first leverage a multimodal dialog simulator (\refsec{sec:dialog_simulator}) to
    generate synthetic dialog flows between a user and an assistant.
    Each of these flows is grounded in a graph connecting the memories of a user from their collection.
    %Note that 
    The memory graphs in our work are simulated by a novel graph simulator (\refsec{sec:graph_simulator}) and 
    are designed to capture several hierarchical relationships between the user memories.
    % leading to to interesting conversational patterns.
}

\begin{figure}[t]
    \centering
    \includegraphics[width=0.99\columnwidth]{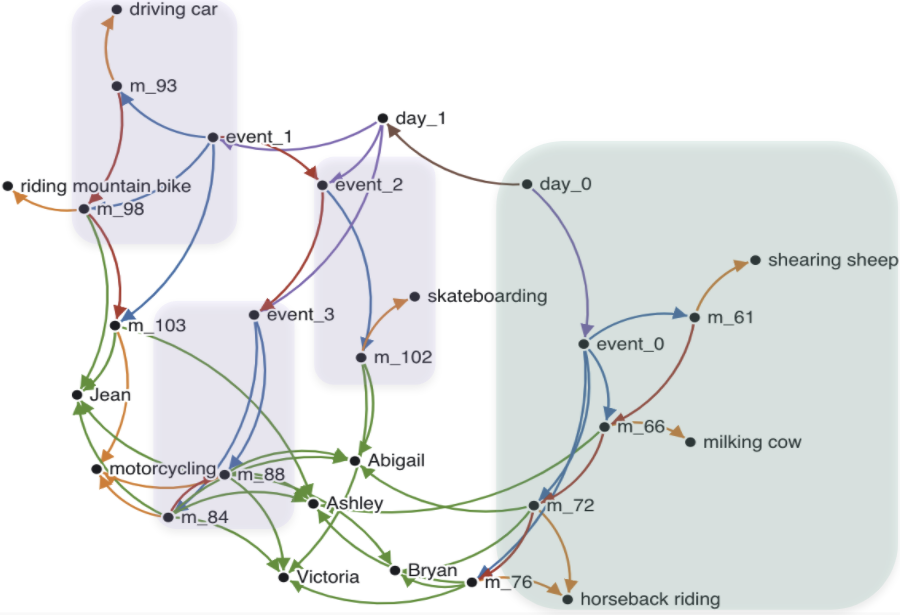}
    \caption{
        Memory subgraph with constituent memories and their hierarchical relationships.
        Each memory contains activity (orange), people (green), time, and place (not shown) attributes.
        Memories are grouped into events (purple box), then days (green box), and finally trips (shown subgraph).
        Each memory graph contains multiple trips, though only one is shown here for brevity.
    }
    \vspace*{\captionReduceBot}
    \label{fig:memory_graph_example}
\end{figure}

\subsubsection{Memory Graph Simulator}
\label{sec:graph_simulator}

\sk{
    Graphs have been ubiquitously used in various fields to effectively represent a set of entities and relationships
    between them.
    Following this trend, we use a graph structure to represent a collection of memories 
    (see \reffig{fig:memory_graph_example} for an example).
    As mentioned in \refsec{sec:introduction}, to circumvent the lack of readily available datasets for personal 
    photo collections and surrounding privacy issues, we construct a novel graph simulator to synthetically generate 
    memories graphs using public datasets.
    These memory graphs are then used as an input to the multimodal dialog simulator to generate dialog flows.
}

\noindent
\textbf{Memories and Attributes.}
\sk{
    Memories constitute the atomic units of the graph simulator, and can cover a wide variety of media including
    photographs, videos, and user-created montages.
    We limit the scope of memories to represent static images in this work, although most components of our proposed
    framework readily extend to the broader definition.
    As photo collection of individuals is sensitive information, we use publicly available image dataset as a proxy to mitigate the risk.
    Specifically, we use Creative Commons images from MS COCO \cite{mscoco} that contains objects and people in 
    everyday contexts as memories.
}

\sk{
    We then assign four attributes to each of the images as follows:
    (a) \textit{Activity}: 
        Each image in MS COCO has $5$ associated captions. We use sentence-BERT \cite{reimers-2019-sentence-bert} 
        to find the closest activity label from the taxonomy of the ActivityNet dataset \cite{caba2015activitynet}, 
        using average text-similarity to the captions.
        To ensure a good representation, we only keep those with at least $20$ memories resulting in about $138$
        labels covering wide variety of activities.
    (b) \textit{Place:} 
        For each activity, we first manually map it to a place type, which then is randomly mapped to an actual place 
        from a manually curated list.
        For instance, \textit{playing frisbee} $\rightarrow$ \textit{park} $\rightarrow$ 
        \textit{Cal Anderson Park, Seattle, USA}.
    (c) \textit{People:} 
        We use the associated bounding box annotations for MS COCO images and map those labeled as `person',
        above a threshold size, to a random name from a curated list of $200$ names.
    (d) \textit{Time}
        attribute is sampled randomly from a constrained time range, depending on the relationship
        %shared by a memory 
        shared with other memories in the graph.
        %The various relationships are elaborated in the following paragraphs.
        %This would be elaborated more in the following paragraph after introducing the various relationships.
}

\noindent
\textbf{Hierarchical Relationships.}
\sk{
    To closely emulate scenarios in a personal photo collection, we devise the following hierarchy of 
    relations amongst the memories:
    \textit{memories} $\rightarrow$ \textit{events} $\rightarrow$ \textit{days} $\rightarrow$ \textit{trips}.
    Using heuristic rules, we sample and group memories into events that are then grouped into days,
    which are finally grouped into trips.
    These groupings impose constraints on the attributes of the constituent memories, which can be used to 
    generate interesting conversational flows to query connected memories.
    For instance, memories from the same event need to happen at the same place type, while those in a day 
    need to happen in the same city.
    Similar restrictions arise for the time attribute as well, which would be used to sample reasonable times
    for the corresponding memories, \eg., memories from the same event cannot be separated by more than few hours.
    These hierarchical relationships enable connected queries like \textit{`What did we do after this?'},
    \textit{`Show other pictures with Jane on this trip'}, or \textit{`Where did we go the next day?'}.
}

\noindent
\textbf{Memory Graphs.}
\sk{
Putting everything together, we construct a memory graph for each collection:
\begin{itemize}[
    noitemsep,topsep=0pt,parsep=0pt,partopsep=0pt,itemindent=0pt,leftmargin=10pt
]
    \item nodes: memory, event, day, trip, person, activity
    \item edges: memory attributes, hierarchical relations
\end{itemize}
Note that each memory graph can contain multiple trips. 
\reffig{fig:memory_graph_example} illustrates a memory subgraph, visualizing only one trip for brevity.
%Using this simulator, we can 
We synthetically generate multiple memory graphs which form the input to the
dialog flow simulator.
}

\noindent
\textbf{Applications in the Real-World Setting}. 
While we use the publicly available image dataset to generate memory graphs, applying the method above in the existing real-world photo album products at large-scale is straightforward as we do not require any additional information (\eg captions or annotations) other than meta data that are readily associated with the media (\eg timestamp, locations).
This meta data can be rearranged from tables to graphs without additional annotations, only by specifying the relations of interest (e.g., people, place, time, predicted concepts). Memory graphs are not only practical but also desired to enable connected memory search.

% Finally, a memory graph can potentially contain multiple trips.

\subsubsection{Multimodal Dialog Simulator}
\label{sec:dialog_simulator}
% \smtodo{paraphrase more}

The multimodal dialog simulator takes the generated memory graphs along with the meta information of each node %(activity, people, locations, time, etc.)
to create user$\leftrightarrow$assistant dialog flows, following the agenda-based dialog simulator approach \cite{schatzmann2007agenda}.

\sm{
\noindent 
\textbf{Dialog Flow Generation via Self-play.}
The dialog simulator comprises three main components: the \textit{goal generator}, the \textit{user simulator}, and the \textit{assistant simulator}.
The goal generator randomly samples an \textit{agenda} for each dialog, which defines a sequence of high-level \textit{goals} for the scenario (\eg, \texttt{SEARCH} $\rightarrow$ \texttt{GET\_RELATED\_PHOTOS} $\rightarrow$ \texttt{GET\_INFO}).
Given a goal, the user simulator draws an acceptable dialog action based on a probability distribution, which is defined with NLU intents (\eg, \texttt{REQUEST:GET}, \texttt{CONFIRM:SHARE}), 
slots (\eg, location, time), and memory references.
The assistant simulator then takes the output of the user simulator, retrieves the multimodal contexts via the simulation API (\eg obtaining the information of a memory node from the graph, retrieving related memories), and generates natural language generation (NLG) intents, slots and new memory references.
The process is repeated until the simulator exhausts every goal in the agenda.
%, or until the maximum number of turns.
}

\sm{
\noindent
\textbf{Multimodal Dialog Ontology.}
Following other task-oriented dialog datasets \cite{multiwoz2.1, sgd-dst, moon2020situated}, for \dn{} we provide the standard dialog annotations such as the intent (NLU \& NLG) and slot labels.
To accommodate for the complex multimodal nature of the scenarios, we extend the dialog ontology to include memory reference annotations as their corresponding node IDs, which seamlessly annotates both multimodal contexts and language (\eg \textit{`When was our trip to Whistler?'} $\rightarrow$ \texttt{INFORM:GET\_INFO.time}, \textit{memories: [8]}). % slots: \{location: Whistler\},
The same notation can be used to refer the memories that are carried over in the dialog context (\eg \textit{`Where did we go after that?'} $\rightarrow$ \texttt{INFORM:GET\_RELATED.location}, \textit{memories: [8]}).
This proposed fine-grained and unified ontology will allow a systematic approach to study diverse referring expressions in multimodal dialogs.  
}

\subsection{Manual Paraphrase}
\label{sec:human_paraphrase}
\sk{
    We then collect manual paraphrases of the generated memory-grounded dialog flows.
    This allows us to draw utterances from the natural language distribution, thus moving closer to the application.
    We build an interactive tool to aid annotators -
    specifically, the interface shows the images corresponding to the memories along with the dialog flow, and
    instructs annotators to paraphrase without losing key information such as objects and attributes.
    The \dn dataset thus comprises many rich visual references, making it an ideal dataset for studying multimodal language grounding.
    See appendix for an example dialog.
    As paraphrasing utterances is faster, cheaper, and requires little to no domain knowledge on the
    annotator`s part, our two-phase pipeline is much more resource-effective, when compared to collecting
    multimodal 
    human$\leftrightarrow$human dialogs and collecting dialog annotations on top \cite{moon2020situated}.
}

\subsection{\dn Dataset Analysis}
\label{sec:dataset_analysis}
\begin{table}[t]
    \begin{center}
        \scalebox{0.78}{
        \begin{tabular}{lccc}
        \toprule[\heavyrulewidth]
        Total \# dialogs & $11.5k$ \\
        Total \# utterances & $103.4k$ \\
        Total \# memory graphs & $1.1k$ \\
        Avg \# words (user turns) & $10.7 \pm 4.4$ \\
        Avg \# words (assistant turns) & $15.4 \pm 9.8$ \\
        Avg \# utterances / dialog 	& $8.8$ \\
        Avg \# memories mentioned / dialog &	$3.5$ \\
        Avg \# memories in graph / dialog &	$100$ \\
        \bottomrule[\heavyrulewidth]
        \end{tabular}
        }
    \end{center}
    %\vspace{-10pt} 
     \vspace*{\captionReduceTop}
    \caption{\textbf{\dn{} Dataset Statistics}}
    \vspace*{\captionReduceBot}
    %\vspace{-10pt}   
    \label{tab:dataset_statistics}
\end{table}

\sk{
%We now analyze the \dn dataset, which contains $11.4k$ dialogs
\dn contains $11.4k$ dialogs
totalling $103.4k$ utterances, grounded in $1.1k$ memory graphs.
\reftab{tab:dataset_statistics} presents the overall dataset statistics.
}

\begin{figure*}[t]
    \centering
    \begin{subfigure}[b]{0.32\textwidth}
        \includegraphics[width=\textwidth]{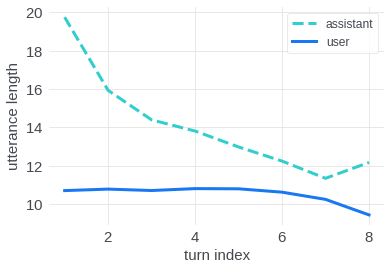}
        \caption{}
        \label{fig:utterance_len_distr}
    \end{subfigure}
    ~
    \begin{subfigure}[b]{0.32\textwidth}
        \includegraphics[width=\textwidth]{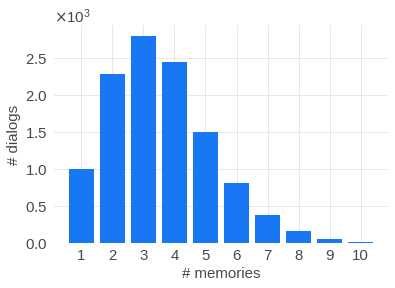}
        \caption{}
        \label{fig:memory_dialog_distr}
    \end{subfigure}
    ~
    \begin{subfigure}[b]{0.32\textwidth}
        \includegraphics[width=\textwidth]{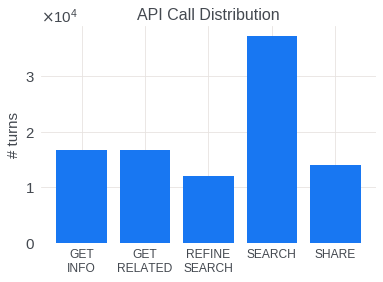}
        \caption{}
        \label{fig:api_distr}
    \end{subfigure}
    
    \begin{subfigure}[b]{0.32\textwidth}
        \includegraphics[width=\textwidth]{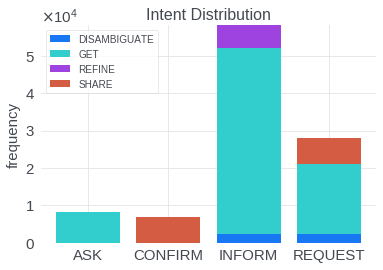}
        \caption{}
        \label{fig:dialog_act_distr}
    \end{subfigure}
    ~
    \begin{subfigure}[b]{0.64\textwidth}
         \includegraphics[width=\textwidth]{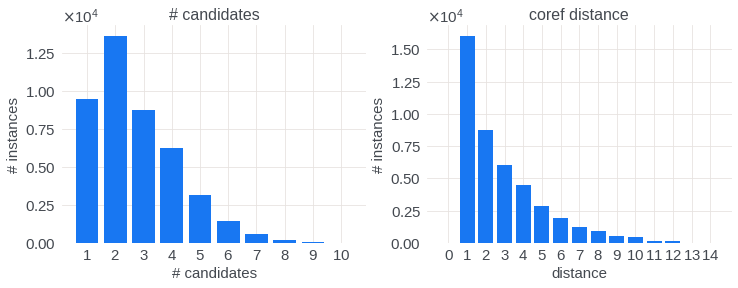}
        \caption{}
        \label{fig:coreference_distance_distr}
    \end{subfigure}
    \vspace*{\captionReduceTop}
    \caption{Distribution of 
    (a) utterance lengths with dialog turns,
    (b) number of memory mentions in each dialog,
    (c) API calls across the dialogs,
    (d) dialog acts and activities, and
    (e) referent candidates (L) and coreference distance (R) between memory mentions.}
    \vspace*{\captionReduceBot}
\end{figure*}

\begin{figure*}
    \centering
    \includegraphics[width=0.95\textwidth]{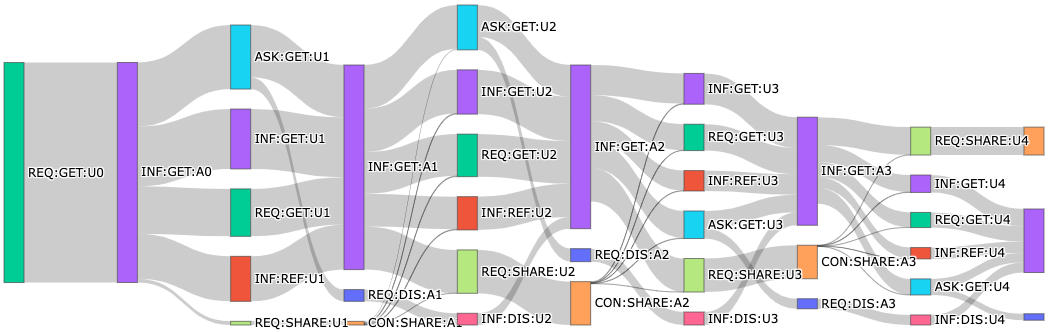}
    \vspace*{\captionReduceTop}
    \caption{
        Transition of dialogs acts in \dn for the first $4$ turns, for dialog flows generated by our
        novel multimodal dialog simulator for connected memories.
        Each block is of the form \texttt{ACT:ACTIVITY:[A|U][turn]}, to denote dialog act, activity, 
        user or assistant turn, and turn number, respectively. See text for more details.
    }
    \vspace*{\captionReduceBot}
    \label{fig:intent_transitions}
\end{figure*}

\noindent
\textbf{Analyzing Dialogs.}
\sk{
    Dialogs in \dn use a total of $1.1k$ memory graphs with each containing $100$ memories.
    For every dialog, there are about $3.5$ connected memory mentions with the distribution
    given in \reffig{fig:memory_dialog_distr}.
    %As the 
    User and assistant turns average about $10.7$ and $15.4$ words respectively 
    (distribution in \reffig{fig:utterance_len_distr}).
    It is interesting to note that the assistant responses are significantly longer than the user. 
    As an example, consider the following user utterance
    \textit{`U: Are there any similar photos from 2020?'} and the corresponding assistant response
    \textit{`A: Here`s one of Laura and Virginia cooking sausages at home, the afternoon of 
    August 26, 2020. It looks like a fun time!'}.
    This illustrates that the annotators paraphrasing the dialog flows included:
    (a) details about the retrieved memories to give additional context to the user, thus invoking
    subsequent connected memory queries (\eg, \textit{`What did we do that evening?'}),
    (b) chitchat about the memories to make the conversational natural sounding.
}

\noindent
\textbf{Analyzing Dialog Annotations.}
\sk{
    Our \dn come with annotations at dialog level for dialog state tracking 
    (NLU intents and slots), necessary API calls for assistant, and multimodal
    coreference resolution.
    Following \citet{simmc2}, our intents follow a hierarchy of \textit{dialog acts} 
    (4: \texttt{ASK}, \texttt{CONFIRM}, \texttt{INFORM}, \texttt{REQUEST}) 
    and \textit{activities} 
    (4: \texttt{DISAMBIGUATE}, \texttt{GET}, \texttt{REFINE}, \texttt{SHARE}).
    See \reffig{fig:dialog_act_distr} for a breakdown distribution over dialog acts and 
    activities.
    Due to the retrieval nature of our assistant (either memories or associated attributes),
    a major chunk of the activities are \texttt{GET}.
    Similarly, there are $5$ APIs in our dataset (\reffig{fig:api_distr}):
       \begin{itemize}[
        noitemsep,topsep=0pt,parsep=0pt,partopsep=0pt,itemindent=0pt,leftmargin=10pt
    ]
        \item \texttt{SEARCH}: Search using input parameters,
        \item \texttt{REFINE\_SEARCH}: Build on top of search carrying over existing parameters,
        \item \texttt{GET\_INFO}: Seek information about current or previouly viewed memories,
        \item \texttt{GET\_RELATED}: Explore other memories similar to the current/prior memories, and,
        \item \texttt{SHARE}: Share it to friends or family,
    \end{itemize}
    % Following closely, the corresponding API calls are shown in \reffig{fig:api_distr}.
    As expected, \texttt{SEARCH} is the most dominant API call in the dataset.
    Note that the turns with \texttt{GET\_RELATED} and \texttt{REFINE\_SEARCH} API calls 
    elevate the need for conversation in retrieving connected memories, where the user requests
    for memories similar to the ones already viewed or with additional specifications,
    respectively.
    Finally, \reffig{fig:coreference_distance_distr} visualizes the distribution of number 
    of candidates and utterance difference between the current and the one with referent 
    memory (coreference distance).
    For turns requiring coreference resolution, the average number of candidates is $2.7$ at a
    distance of $2.9$ utterances.
    Though a majority of referents are naturally $1$ utterance away (previous turn), the long
    tail (even up to $10+$ utterances) indicates the presence of challenging multimodal 
    coreferences.
}

\noindent
\textbf{Analyzing Dialog Flows.}
\sk{
    As mentioned earlier, the multimodal dialog simulator generates the dialog flows during the first phase of our
    data generation.
    We visualize these dialogs flows in \reffig{fig:intent_transitions} for the first four dialog turns,
    where each block denotes an intent at a particular turn and the grey stripes denote NLU intent transitions
    in subsequent turns.
    The width of the stripe is proportional to the frequency of the transition.
    For brevity, each block is label as \texttt{ACT:ACTIVITY:[A|U][turn]}.
    %As remarked earlier, \texttt{INFORM:GET} is indeed the dominant assistant intent due to the nature of the problem.
    The high branch-off factors for these intents capture the diversity of the dialogs flows in our dataset,
    which is desirable in building a robust dialog system.
}

\section{Task Formulation}
\label{sec:task_formulation}
\begin{table*}[t]
    \begin{center}
        \scalebox{0.69}{
        \setlength\tabcolsep{9pt}
        \begin{tabular}
        {p{0.35\textwidth} p{0.58\textwidth} p{0.33\textwidth}}
        
        \toprule[\heavyrulewidth]
        \textbf{Task Name}  & \textbf{Goal} & \textbf{Evaluation} \\
        \midrule
        %1. Multimodal Disambiguation & \todo{fill} & Ambiguity Detection Precision / Recall / F1 \\
        %\midrule      
        
        1. Assistant API Call Prediction
        & Given user utterances, predict the right API call necessary to execute the query.
        & Classification accuracy \\
        
        \midrule
        
        2. Multimodal Coreference \newline Resolution (MM-Coref) & 
        Given user utterances, resolve referent memories to their
        canonical ID(s) as defined by the memory graph. 
        & Coref Precision / Recall / F1 \\
        
        \midrule
        
        3. Multimodal Dialog State Tracking \newline (MM-DST)
        & Given user utterances, track user belief states across multiple turns.
        & Slot Precision / Recall / F1 \\
        
        \midrule
        
        4. Assistant Response Generation 
        & Given user utterances, ground-truth APIs and ground-truth object IDs, 
        generate Assistant responses or retrieve from a candidate pool.
        & Generation: BLEU;
        
        Retrieval: Accuracy@k, mean reciprocal rank, mean rank \\
        \bottomrule[\heavyrulewidth]
        \end{tabular}
        }
    \end{center}
    \vspace*{\captionReduceTop}
    \caption{
    Proposed tasks and descriptions on our \dn{} dataset. 
    Please see \refsec{sec:task_formulation} for more details.
    }
    %\vspace*{\captionReduceBot}
    \vspace{-5pt}
    \label{tab:tracks}
\end{table*}

% Task Formulation
% API call prediction
% Dialog State Tracking
% Response Generation
% Coreference resolution

% GET_INFO, GET_MEMORY, REFINE_MEMORY, GET_RELATED, SHARE
% (Tentative) Memory Retrieval

\sk{
To benchmark progress of conversational models towards the goal of assisting users in interactively querying
connected memories in a meaningful way, we propose four main tasks for \dn.
\reftab{tab:tracks} outlines the tasks and the evaluation metrics.
%\reftab{tab:tracks} outlines the task formulations along with the corresponding evaluation metrics.
%We now present the motivation for each of these tasks and dive deeper into the details.
}

\subsection{Assistant API Call Prediction}
\sk{
    The first step in executing any query on connected memories successfully is to
    understand the user utterance in the context of the dialog history and multimodal
    information, and predict the right API call.
    For instance, a query like \textit{`When was the last time I was here?'} should 
    result in a \texttt{GET\_INFO} API prediction.
    Note that errors in API call prediction cascade through the model pipeline resulting 
    in an incorrect or unrelated response from the assistant.
    Thus, this task tests the ability of the conversational agent to predict the right 
    API call. Evaluation is done per each turn through API call accuracy.
}

\subsection{Multimodal Coreference Resolution}
\sk{
    Recall that one of our motivations to use conversations for querying connected memories
    is the ability to support multi-turn queries.
    In such scenarios, humans often use short-hands or references when the underlying 
    referred entity (referent) can be usually deduced without any ambiguity.
    As an example, when looking at a particular memory, a follow-up
    \textit{`When was the last time I was \underline{here}?'} 
    is intuitive and natural, whereas 
    \textit{`When was the last time I was at \underline{Waikiki Beach, Hawaii}?'}
    requires the user to remember the name and use it in the query, making it cumbersome.
}

\sk{
    Therefore, the model must be able to handle multimodal coreferences in order to field
    such queries effectively.
    The input for this task includes the dialog history, multimodal context, and all the
    memories mentioned so far (as coreference candidates).
    The models needs to thus resolve the reference to one or more of the candidates.
    We use coreference precision, recall, and F1 to measure performance.
}

\subsection{Multimodal Dialog State Tracking}

\sk{
    Due to the multimodal nature of \dn, we adopt multimodal dialog state tracking (MM-DST) 
    used in \cite{simmc2} as one of our tasks.
    To elaborate, slots in our dataset can be grounded in the multimodal context information and
    requires reasoning through the current or previously viewed memories.
    For instance, a query like \textit{`Where did we go from here?'} requires the slot value to
    be the currently viewing memory.
    This implies that the dialog states can contain non-textual tokens (\eg, memories), thus
    making it multimodal.
    In order to measure the performance in this task, we use slot recall, precision, and F1 
    scores.
    Note that unlike \cite{simmc2}, we drop evaluating for dialog act prediction since
    \texttt{GET} has an overwhelming majority due to the nature of the problem.
}

\subsection{Assistant Response Generation}
\sk{
    %This task focuses on evaluating 
    This task evaluates the ability of the model to either generate a response or retrieve from a pool of 
    candidates, given dialog history, ground-truth APIs \& results, belief states, and multimodal contexts.
    Though the model has access to API results, producing a natural language utterance to describe it within
    the flow of the dialog is still a difficult task.
}

\sk{
%Following prior work \cite{simmc2},
We evaluate this task in two different ways:
(a) \textit{Generative}, where the model produces the response similar to a conditional language model. 
We use n-gram overlap based BLEU-4 \cite{papineni-etal-2002-bleu} and 
more recent neural evaluation metric BERTScore \cite{bert-score} to measure
performance by comparing the generated response to the ground truth, and
(b) \textit{Retrieval}, where the model ranks a list of randomly pooled candidate responses (unique to a turn)
along with the ground truth. Retrieval metrics like recall@k ($k=\{1, 5, 10\}$), mean rank, and mean reciprocal rank
are used. % to compare model performances
}

\section{Modeling \& Empirical Analysis}
\label{sec:model_analysis}
\sk{
    We now perform preliminary empirical evaluation and analysis for the proposed tasks by training baselines.
    Detailed modeling work is left as future work.
}
\begin{table}[t]
    \setlength{\tabcolsep}{2pt}
    \begin{center}
        \scalebox{0.72}{%
            \begin{tabular}{
                ccccccc
            }
            \toprule[\heavyrulewidth]
            \multirow{2}{*}{\textbf{Model}}
            & \textbf{1. API}
            & \textbf{2. Coref}
            & \multicolumn{2}{c}{\textbf{3. DST.}}
            & \multicolumn{2}{c}{\textbf{4. Gen.}} \\
            % & \textbf{}\\
            \cmidrule(r){2-2}
            \cmidrule(r){3-3}
            \cmidrule(r){4-5}
            \cmidrule(r){6-7}
                &
                Acc$\uparrow$
                & Coref F1$\uparrow$
                & Slot F1$\uparrow$
                & Joint Acc.$\uparrow$
                & BLEU$\uparrow$ &
                BERTS.$\uparrow$ \\
            \midrule
            Text
                & \reportval{88.6}{0.3}
                & \reportval{78.2}{0.4}
                & \reportval{91.5}{0.4}
                & $72.9$
                & $0.205$ %\reportval{0.385}{0.004}
                & $0.895$ \\ %\reportval{0.884}{0.001} \\
            \midrule
            MM-{\small BUTD}
                & \reportval{89.4}{0.4}
                & \reportval{84.8}{0.6}
                & \reportval{\textbf{92.6}}{0.3}
                & $77.5$ 
                & $\textbf{0.251}$ %\reportval{\textbf{0.391}}{0.004}
                & $\textbf{0.905}$ \\%\reportval{\textbf{0.904}}{0.001}\\
            MM-{\small CLIP}
                & \reportval{\textbf{90.2}}{0.1}
                & \reportval{\textbf{84.9}}{0.4}
                & \reportval{\textbf{92.6}}{0.2}
                & $\textbf{78.3}$
                & $\textbf{0.251}$
                & $\textbf{0.905}$ \\%\reportval{0.901}{0.001}\\
            \bottomrule[\heavyrulewidth]
            \end{tabular}
        }
    \end{center}
    %\vspace*{-10pt}
    \vspace*{\captionReduceTop}
     \caption{
        Baseline performances for GPT-2 models: text-only (text) and multimodal image features (MM).
        \textbf{(1) API Call Prediction (API)}, via 
        \underline{acc}uracy, 
        \textbf{(2) Multimodal Coreference Resolution (Coref)}, via \underline{coref} prediction \underline{F1},
        \textbf{(3) Dialog State Tracking (DST)}, via \underline{slot F1},
        \textbf{(4) Response Generation} via \underline{BLEU} and \underline{BERTS}core.
        $\uparrow$: higher is better.
        \textbf{Bold}: best performance with statistical significance.
        BLEU and BERTScores have $0.004$ and $0.001$ as stderr values
        respectively.
        %, $\downarrow$: lower is better.
        }
    \vspace*{-4pt}
    %\vspace{-10pt}
    %\vspace*{\captionReduceBot}
    \label{tab:results}    
\end{table}

\noindent
\textbf{Dataset Split.}
\sk{
    The dataset is randomly divided into: train ($70\%$), val ($15\%$), and test ($15\%$).
    For our experiments, models are trained using train split and performance is reported on test, while val is used 
    to pick the model hyper-parameters.
}

\noindent
\textbf{Notations.}
\sk{
    We follow the notation established in \cite{simmc2}, where each dialog of length $N_r$ rounds is represented as 
    $\mathcal{D} = \{(U_i, A_i, M_i, B_i)\}_{i=1}^{N_r}$ with:
    \begin{itemize}[
        noitemsep,topsep=0pt,parsep=0pt,partopsep=0pt,itemindent=0pt,leftmargin=10pt
    ]
        \item $U_i$: User utterance at turn $i$
        \item $A_i$: Assistant utterance at turn $i$
        \item $M_i$: Multimodal context, \ie, memory graph and memories retrieved in the previous turns,
        \item $B_i$: Multimodal belief state, a semantic parse of $U_i$ (intent, slot, memory references). % $U_i$
    \end{itemize}
    Therefore, given the current user utterance ($U_t$), dialog history $H_t = (U_i, A_i)_{i=1}^{t-1}$, and the
    multimodal context ($M_t$), a \dn agent should predict the user belief state $B_t$ and the natural language
    response $A_t$ for every dialog turn $t$.
}

\begin{figure}
    \centering
    \includegraphics[width=0.99\columnwidth]{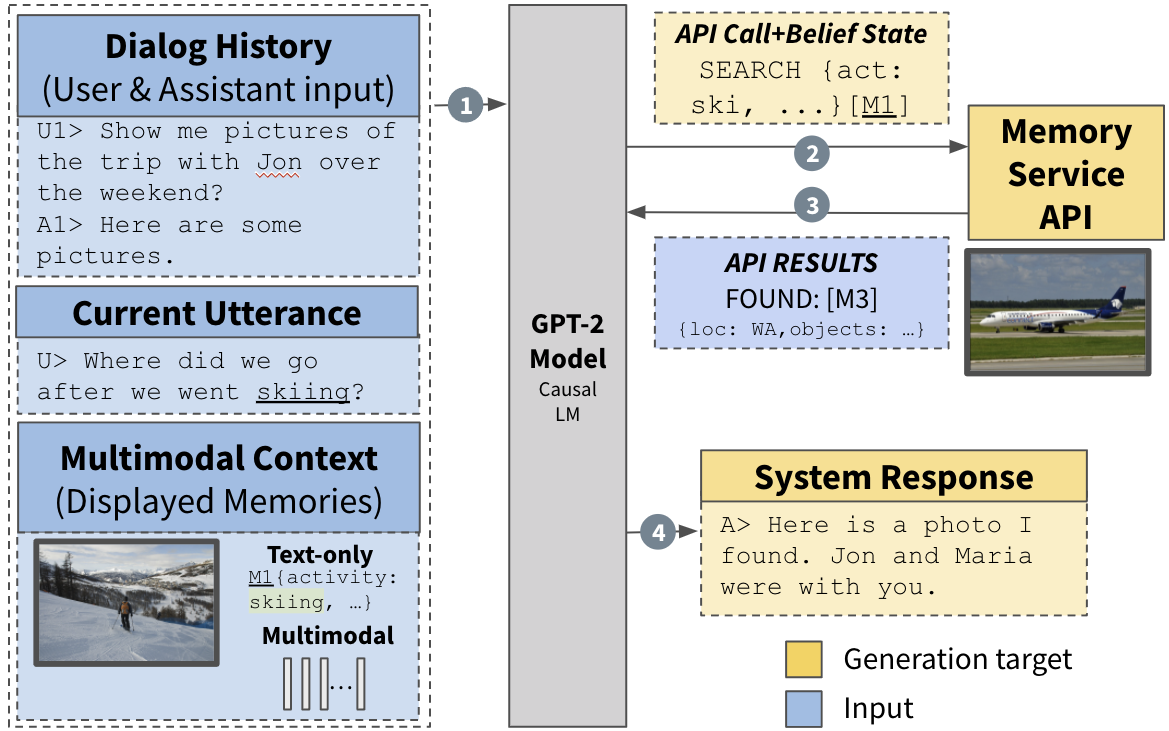}
    \caption{
        Baseline GPT-2 models for \dn. 
        (1) Given the dialog history, multimodal context, and current user utterance, the model
        predicts the API call and belief state at the current turn,
        (2) The API call is executed and (3) the results are fed back into the model,
        (4) Finally, model produces a natural language response.
        As shown, GPT2-text uses attribute strings to represent memories, while GPT2-MM use image features.
    }
    \vspace*{\captionReduceBot} 
    \label{fig:baseline_model}
\end{figure}

\noindent
\textbf{Baselines.}
\sk{
    Causal language models pretrained on large datasets have shown a lot of promise in multimodal and text-only
    task-oriented dialog modeling, when finetuned on the downstream task 
    \cite{simpletod,soloist,simmc2,moon2020situated}.
    Following this popular approach, we adopt the transformer-based GPT-2 \cite{radford2019language} model and 
    jointly train it for API prediction, MM-Coref, DST, and response generation tasks, as shown in 
    \reffig{fig:baseline_model}.
    In particular, we use the 12-layer GPT-2 ($117M$) model and finetune it on dialogs from \dn dataset, using
    early stopping based on token perplexity (<3 GPU hrs).
    % on the val split 
    We use two approaches to capture $M_i$: \newline
    (a) \textit{text-only} (GPT2-text), where previously viewed memories and their attributes are represented as 
        flattened strings. Note that this baseline uses ground-truth activities from the memory graph. \newline
    (b) \textit{multimodal} (GPT2-MM), where bottom-up and top-down ({\small BUTD}) 
    \cite{Anderson2017up-down} and {\small CLIP} \cite{radford2021learning} image 
        features are extracted for previous viewed memories, and fed as `visual tokens'
        while finetuning the GPT-2 model.
}

\noindent
\textbf{Analysis.}
\sk{
    % \reftab{tab:results} summarizes the performance of our baselines on the four proposed tasks.
    A key observation from \reftab{tab:results} is that multimodal models outperforms text-only across all the tasks significantly. 
    This is intuitive, for instance, multimodal coreference resolution requires understanding the memories beyond the obvious
    activity label in order to rightly resolve the reference.
    Consider the query: \textit{`When was the last time I played with my dog \underline{here}?'}.
    To resolve to the right memory, the system needs to understand which memory is about playing with the dog
    towards which a mere activity label \textit{throwing frisbee} might be insufficient.
    For a similar reason, additional multimodal features improve response generation, especially to include chit-chat.
    On the other side, GPT-Text performs competitively on capturing the dialog state.
}

% \section{Conclusion}
% \label{sec:conclusion}
\noindent \textbf{Conclusion.}
\sk{
We present a novel dataset for the dialogs for connected memories,
\dn{}, with $11.5K$ user$\leftrightarrow$assistant dialogs ($103K$ utterances) grounded on the memory graphs.
We present a novel multimodal dialog simulator, which generates simulated dialogs grounded on diverse memory graphs that are automatically configured.
Our empirical analysis demonstrates many new challenges that our \dn{} dataset brings, 
highlighting new directions of research in this area.
%Next, we discuss some limitations of our work and related ethical considerations.
}

% --------------------------------------------
% https://2022.emnlp.org/calls/main_conference_papers/
% Mandatory Discussion of Limitations
% We believe that it is also important to discuss the limitations of your work, in addition to its strengths. EMNLP 2022 requires all papers to have a clear discussion of limitations, in a dedicated section titled “Limitations”. This section will appear at the end of the paper, after the discussion/conclusions section and before the references, and will not count towards the page limit. Papers without a limitation section will be automatically rejected without review.
% --------------------------------------------

\noindent
\section{Limitations}
\sk{
The generalizability of \dn is naturally bounded by the underlying graph simulator,
    %Given the heuristic nature of the graph simulator
    especially around memory attribute labels of place,
    people, and time.
    %its generalizability is naturally limited.
    % In other words, the arbitrary selection of places/names lists in the simulator has an impact on the 
    % robustness of the downstream dialog model.
    However, we justify this as follows:
    (a) Recall that the focus of our work is to enable an assistant that can understand and execute user 
    queries about connected memories through an interactive dialog.
    %Even with our somewhat limiting simulator, 
    Even with the simulated dialog flows, \dn captures several interesting
    challenges related to multimodal dialog, for instance, coreference resolution and dialog state tracking 
    (as seen in \refsec{sec:dataset_analysis} and \refsec{sec:model_analysis}).
    This opens the door to new research directions in multimodal conversation, especially in the absence 
    of a readily available large-scale personal photo collection dataset (along with attributes and metadata).
    (b) Due to the two-stage data collection pipeline, \dn is amenable to data augmentation techniques 
    that can increase the robustness of the downstream dialog model.
    For instance, the dataset can be easily augmented by replacing named entities in the memory graph 
    and utterances, without changing the flow.
}

\emptydraft{
\noindent
\textbf{Ethical Considerations.} All identifiable faces from the COCO images are blurred using a CV algorithm, mitigating privacy risks. Annotators for our task were employed as full-time and contracted via a leading NLP/linguistics annotation platform.
% \sk{
% As 
% }
}

\bibliography{bibliography}
% %\bibliographystyle{plainnat}
 \bibliographystyle{acl_natbib}

%\vfill
%\pagebreak
\appendix

.
\vfill
\pagebreak
\section{Supplementary Materials}

% --------------------------------------------
% https://2022.emnlp.org/calls/main_conference_papers/
% Mandatory Discussion of Limitations
% We believe that it is also important to discuss the limitations of your work, in addition to its strengths. EMNLP 2022 requires all papers to have a clear discussion of limitations, in a dedicated section titled “Limitations”. This section will appear at the end of the paper, after the discussion/conclusions section and before the references, and will not count towards the page limit. Papers without a limitation section will be automatically rejected without review.
% --------------------------------------------

\emptydraft{
\subsection{Limitations of the proposed work}

%\noindent
%\textbf{Limitations.}
\sk{
The generalizability of \dn is naturally bounded by the underlying graph simulator,
    %Given the heuristic nature of the graph simulator
    especially around memory attribute labels of place,
    people, and time.
    %its generalizability is naturally limited.
    % In other words, the arbitrary selection of places/names lists in the simulator has an impact on the 
    % robustness of the downstream dialog model.
    However, we justify this as follows:
    (a) Recall that the focus of our work is to enable an assistant that can understand and execute user 
    queries about connected memories through an interactive dialog.
    %Even with our somewhat limiting simulator, 
    Even with the simulated dialog flows, \dn captures several interesting
    challenges related to multimodal dialog, for instance, coreference resolution and dialog state tracking 
    (as seen in \refsec{sec:dataset_analysis} and \refsec{sec:model_analysis}).
    This opens the door to new research directions in multimodal conversation, especially in the absence 
    of a readily available large-scale personal photo collection dataset (along with attributes and metadata).
    (b) Due to the two-stage data collection pipeline, \dn is amenable to data augmentation techniques 
    that can increase the robustness of the downstream dialog model.
    For instance, the dataset can be easily augmented by replacing named entities in the memory graph 
    and utterances, without changing the flow.
}
}

\subsection{Ethical Considerations}

%\noindent
%\textbf{Ethical Considerations.}
All identifiable faces from the COCO images were blurred using a CV algorithm, mitigating potential privacy risks. The dataset, when released publicly, will include those edited images.

\begin{figure*}
    \centering
    \includegraphics[width=1.75\columnwidth]{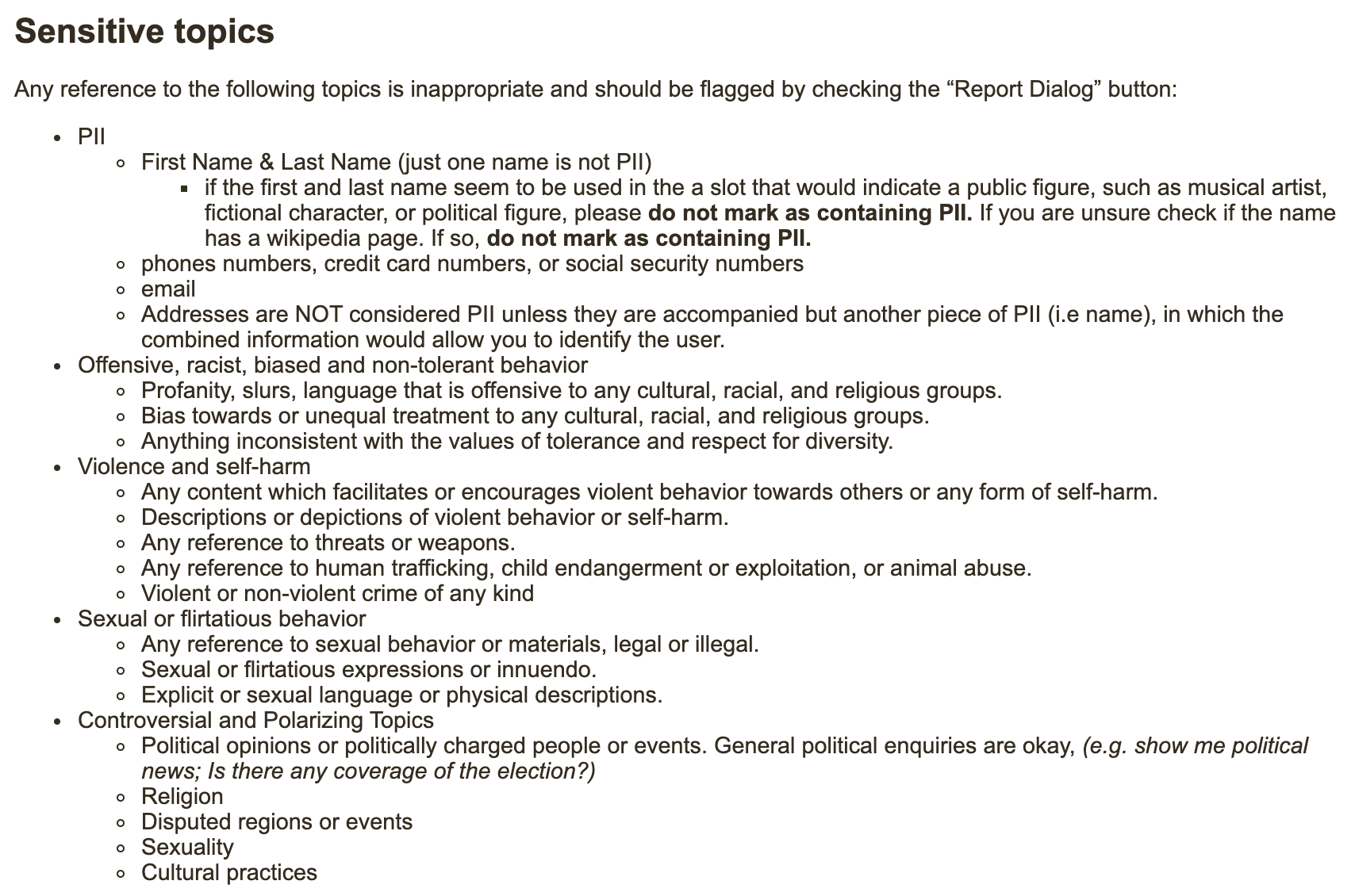}
    \caption{
        Disclaimers shown to the annotators, detailing the escalation path.
    }
    \vspace*{\captionReduceBot} 
    \label{fig:annotaiton_ui_disclaimer}
\end{figure*}

Annotators for our task were employed as full-time and contracted via a leading NLP / linguistics annotation platform.
Annotators were given clear instructions and disclaimers detailing the escalation path (``Report Dialog") for an (unlikely) case where the data may include sensitive topics or images (shown in Figure \ref{fig:annotaiton_ui_disclaimer}).

% \sk{
% As 
% }

\subsection{Dataset Example}

\begin{figure*}[b]
%\begin{figure*}[hbtp]
%\begin{minipage}{\textwidth}
  \centering \includegraphics[width=1.95\columnwidth]{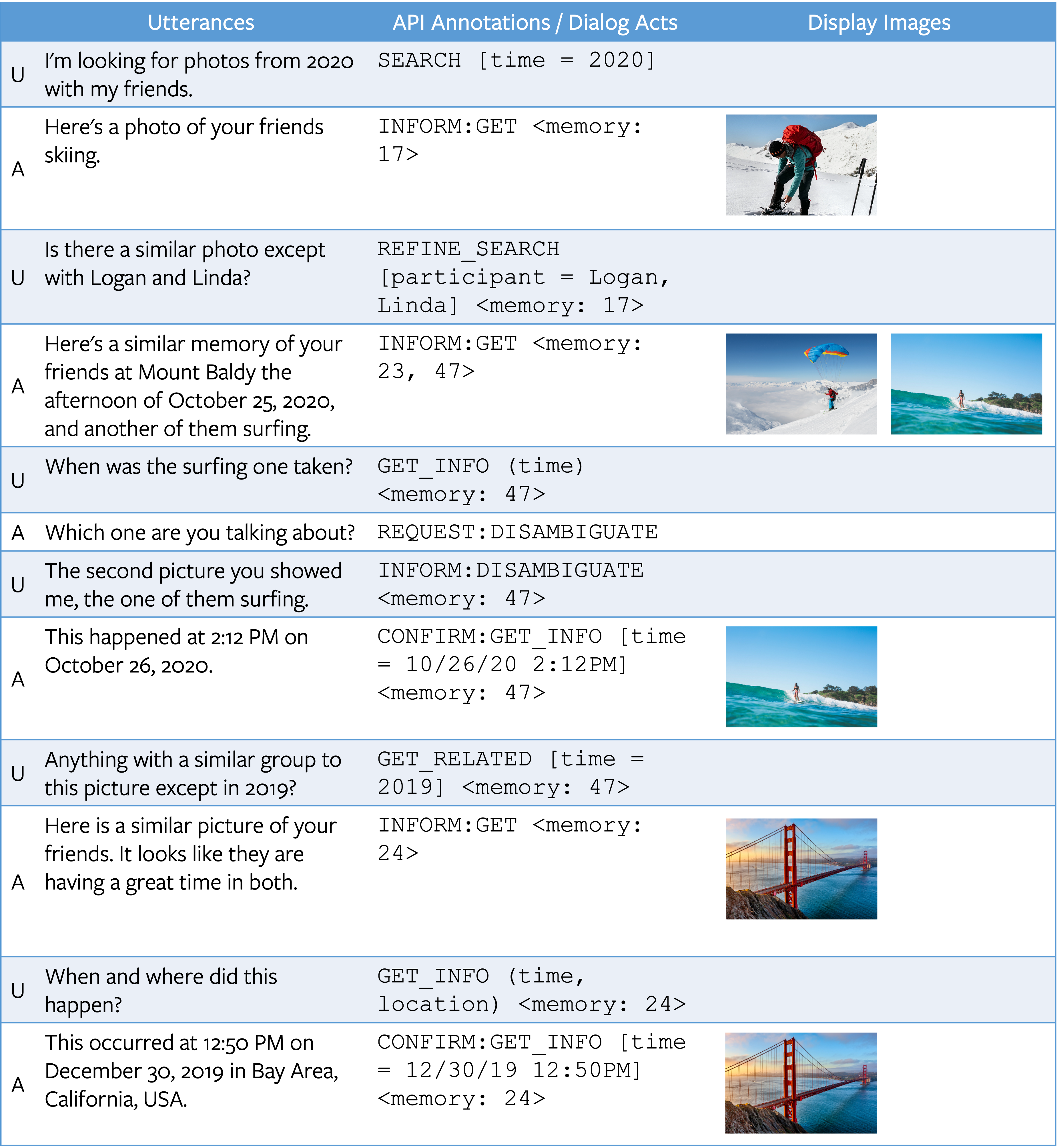}
  %\captionsetup{type=figure}
  %\vspace{-0pt}
    \caption{\textbf{Dataset Example}. Dialog labels include intent, slots, and multimodal coreferences. }
  \label{fig:example}
\end{figure*}

Figure \ref{fig:example} illustrates an example dialog from \dn, along with the set of images associated with each turn (U: User, A: Assistant).
API Annotations are formatted as follows: \texttt{INTENT [slot = value, ...] (request\_slot) <memory: ID>}.
When there is no new image introduced for a given turn, it is assumed that the images from previous turns (if any) are left visible to the user, therefore continuing to serve as the grounding multimodal context. 

%\end{minipage}

\vfill
\pagebreak

%\textbf{Appendix B. Annotation Tool Screenshots}

\subsection{Annotation UI}

\begin{figure*}
    \centering
    \includegraphics[width=1.75\columnwidth]{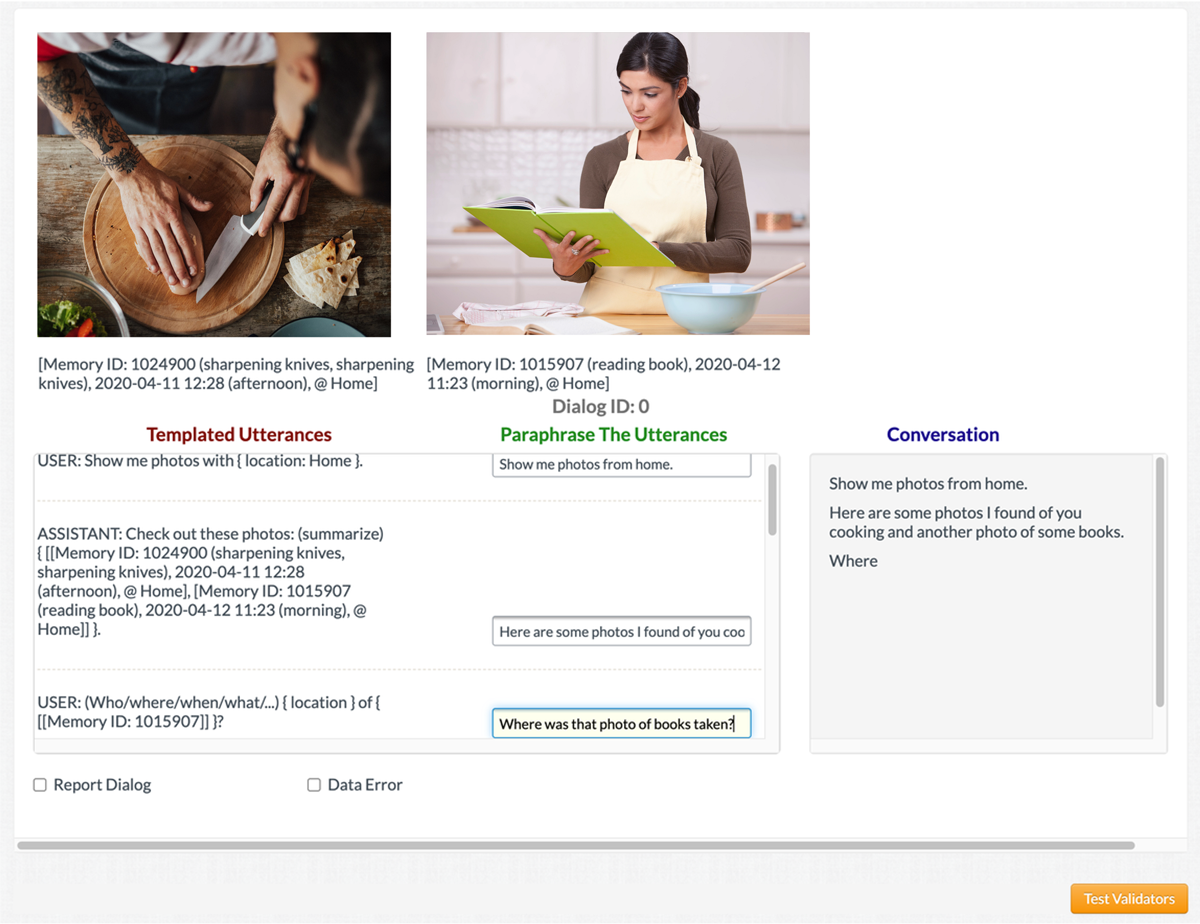}
    \caption{
        \textbf{The annotation tool UI}. Annotators are shown the templated utterances, and a set of photos that dynamically get updated for each turn, based on the pre-generated dialog flows.
    }
    \vspace*{\captionReduceBot} 
    \label{fig:annotaiton_ui}
\end{figure*}

Figure \ref{fig:annotaiton_ui} illustrates the annotation UI used to collect multimodal paraphrases of the dialog.
Annotators are shown the pre-generated dialog flows (templated utterances), along with the text boxes where the paraphrases can be entered. 
The top portion of the UI displays the images (assumed to be) shown to the user for each given turn, which gets dynamically updated as annotators click on new text boxes for entering paraphrases. 
A shortened list of meta data associated with each image is also shown for reference.

%  \newpage
%  \input{sections/author_response.tex}

\end{document}